\documentclass[10pt,conference,a4paper]{IEEEtran}
\usepackage{mathrsfs}
\usepackage{algorithm}
\usepackage{algpseudocode}
\usepackage{mathrsfs}
\usepackage{bm}
\usepackage{stmaryrd}
\usepackage{graphicx}
\usepackage{subfigure}
\usepackage{amssymb}
\usepackage{amsmath}
\usepackage{verbatim}
\usepackage{booktabs}
\usepackage{ctable}
\usepackage{multirow}
\usepackage{multicol}
\usepackage{rotating}
\usepackage{threeparttable}
\usepackage{cite}
\usepackage{collref}
\usepackage{url}
\usepackage[colorlinks=true,linkcolor=red,citecolor=blue]{hyperref}

\makeatletter

\newcommand{\Rmnum}[1]{\expandafter\@slowromancap\romannumeral #1@}
\makeatother

\title{Texture Synthesis Through Convolutional Neural Networks and Spectrum Constraints}
\author{\IEEEauthorblockN{Gang Liu, Yann Gousseau}
\IEEEauthorblockA{{\em Telecom-ParisTech, LTCI CNRS}\\
46 Rue Barrault, 75013 Paris, France.\\
\{{\em gang.liu, gousseau}\}@telecom-paristech.fr}
\and
\IEEEauthorblockN{Gui-Song~Xia}
\IEEEauthorblockA{{\em State Key Lab. LIESMARS, Wuhan University}\\
129 Luoyu Road, Wuhan 430079, China.\\
{\em guisong.xia}@whu.edu.cn}}

\begin{document}
\maketitle

\begin{abstract}
This paper presents a significant improvement for the synthesis of texture images using convolutional neural networks (CNNs), making use of constraints on the Fourier spectrum of the results. More precisely, the texture synthesis is regarded as a constrained optimization problem, with constraints conditioning both the Fourier spectrum and statistical features learned by CNNs. In contrast with existing methods, the presented method inherits from previous CNN approaches the ability to depict local structures and fine scale details, and at the same time yields coherent large scale structures, even in the case of quasi-periodic images. This is done at no extra computational cost. Synthesis experiments on various images show a clear improvement compared to a recent state-of-the art method relying on CNN constraints only.
\end{abstract}

\section{Introduction}

A large body of works has been dedicated over the last 30 years to the task of texture synthesis and more precisely to the task of generating new but perceptually similar images from a given exemplar. Parametric Markov models were among the first to be investigated for such tasks\cite{cross1983markov}. Then, many methods were developed relying on the idea, inspired by the works of Julesz~\cite{julesz1962visual}, that realistic textures could be obtained by constraining a well chosen set of image statistics: wavelet marginals~\cite{heeger1995pyramid}, joint statistics of wavelet coefficients~\cite{Portilla2000}, Fourier spectrum~\cite{galerne2011random,XiaFPA14} or the distribution of sparse representation coefficients~\cite{Tartavel2015}. A completely different approach is provided by patch-based, non-parametric methods that have been largely studied after the initial works of~\cite{Efros1999} and \cite{Wei2000}. Here, the basic idea is to generate a new texture by iteratively sampling patches from the exemplar, see e.g.~\cite{Lefebvre2005} for a review. These methods enable realistic results even on highly structured textures, at the price of very limited innovation capacity, see e.g.~\cite{aguerrebere2013exemplar}. Recently, the work from~\cite{raad2015conditional} renewed this type of methods by promoting the use of parametric distributions for patches.

\begin{figure}[t!]
	\centering
	\includegraphics[width=0.75\linewidth]{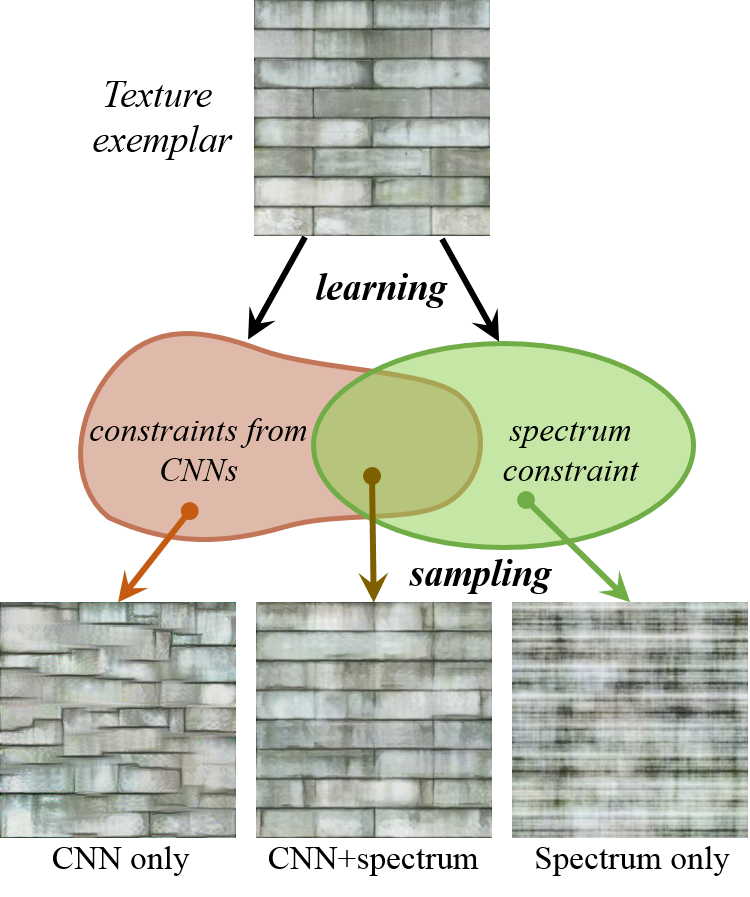}
	\caption{Illustration of the set of constraints used in this work. We impose both the correlations between filter responses in a CNN (in red) and the power spectrum (in green) of the solution. Observe that the synthesis results using only CNN constraints have good local structure and appearance but may lose large scale regularity, whose reproduction is allowed by constraining the spectrum.}
	\label{fig:combinedmethod}
\end{figure}



Recently, several synthesis methods based on convolutional neural networks (CNNs) have been proposed and shown to yield state-of-the art results~\cite{Dai2014,gatys2015texture,Lu2016,Ulyanov2016}. Indeed, these methods are able to generate perceptually satisfying results on complex textures without producing piecewise recopy of the exemplar. The idea, in order to generate new textures, is to constrain some statistics of convolutional networks initially introduced for image classification. In~\cite{gatys2015texture,Ulyanov2016}, a non-parametric approach is proposed through the correlation matrices of filter responses (named \emph{feature maps}), as we will see in more details later. In~\cite{Dai2014, Lu2016}, a MRF model is learned from the feature maps. One limitation of these approaches, however, is the difficulty to efficiently account for large scale regularity, see e.g. Fig.~\ref{fig:combinedmethod} for an illustration.

In this work, we propose to incorporate some low frequency constraints into the CNN approach, in order to allow the synthesis of textures having large scale regularity. In particular, this will allow the synthesis of quasi-periodic textures such as \emph{brick wall} or \emph{object alignments}. In order to do this, we draw on previous works on texture synthesis~\cite{galerne2011random, galerne2011micro} that have shown the interest and easiness of using spectrum constraints for synthesizing textures. We therefore combine spectrum and feature maps constraint through the definition of a new loss function, allowing the reproduction of both fine scale details and large scale regular structures. The combination of these two types of constraints is illustrated in Fig.~\ref{fig:combinedmethod}.


The remainder of the paper is organized as follows. In Sec.\ref{sec:syncnn} we briefly recall how to use CNNs to synthesize textures. We then explain how to incorporate frequency control into this framework in Sec.\ref{sec:synspe}. Eventually, we experimentally show the interest of the approach in Sec.\ref{sec:exp}.


\section{Texture synthesis by CNNs}
\label{sec:syncnn}

\subsection{Convolution neural network}
 A convolution neural network (CNN) is a feed-forward artificial neural network whose neurons process overlapping regions of the input, generally an image. Parameters of such networks are usually learned using a back-propagated algorithm from a large number of annotated inputs. The network VGG-19, proposed in~\cite{simonyan2014very}, is one of the CNN trained with the  ImageNet dataset and has proven to be efficient in describing texture images with only convolution layers~\cite{cimpoi2015deep}. Neglecting the latter fully connected layers, the network contains $16$ linearly rectified convolution layers and $5$ pooling layers. Notice that for the task of texture synthesis, according to~\cite{gatys2015texture}, the max-pooling strategy may be replaced by average-pooling for improving the gradient flow and obtaining slightly cleaner results.


\subsection{Texture model}

In order to use CNNs for texture synthesis, it was first proposed in~\cite{gatys2015texture} to constrain the statistics of feature maps. Inspired by Portilla and Simoncelli~\cite{Portilla2000}, important statistics are provided by correlations between the feature maps corresponding to different filters.

Given a texture exemplar $I\in \mathbb{R}^{N}$, where $N$ is the number of pixels in the image, we first input $I$ to the CNN to calculate the feature maps at each convolution layer. Suppose there are $m_l$ neurons at $l$-th convolution layer, the extracted feature map is denoted by $f^l\in \mathbb{R}^{m_l\times N_l}$, with $N_l$ as the number of stimuli. Note that at the first convolution layer, $N_l = N$.

After obtaining the feature map $f^l$ at the $l$-th layer, the correlation matrix $G^l\in \mathbb{R}^{m_l\times m_l}$~\cite{gatys2015texture} is computed as
\begin{align}
\label{eq:gram}
    G^l_{p,q} =  \sum_{i=1}^{N_l} f^l_p(i)\cdot f^l_q(i),
\end{align}
where $p,q$ denote the index of feature map corresponding to the filter $p, q\in \{1,\cdots,m_l\}$.

The set of correlation matrices $\{ G^1, G^2, \cdots, G^L\}$ from the different layers are used to model the texture~\cite{gatys2015texture} and to constrain the synthesis of new texture images.

\begin{figure*}[htb!]
\centering
  \includegraphics[width=0.8\linewidth]{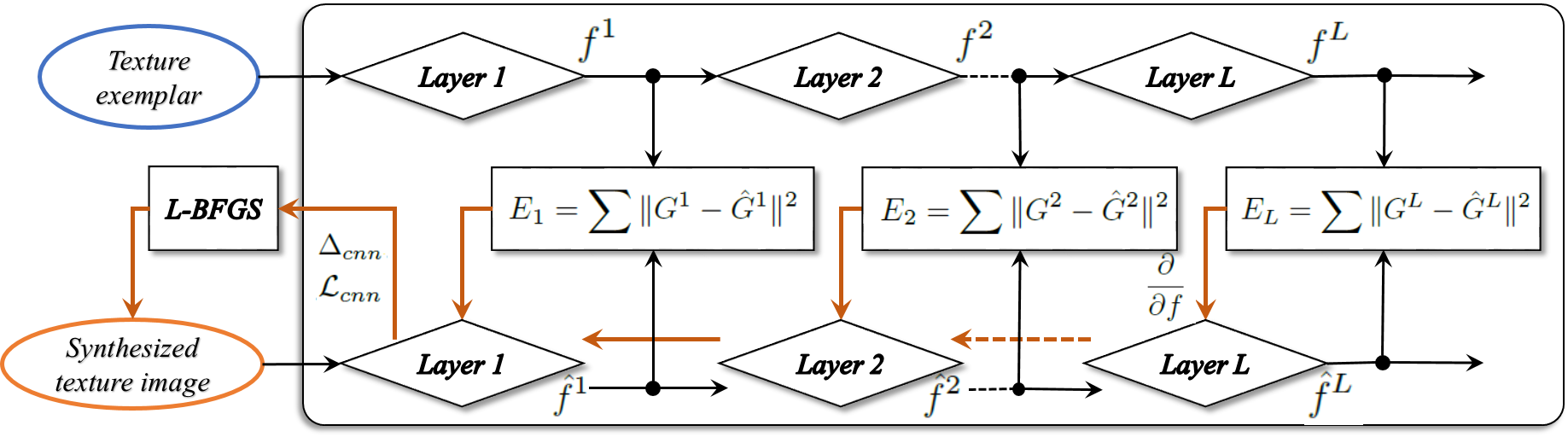}
\caption{The exemplar-based texture synthesis algorithm with CNN model~\cite{gatys2015texture}. At each layer of the convolution neural network, the feature maps $f^l,\hat{f}^l$ are extracted for both the reference image and the generated image. Then the errors $E^l$ between the correlation matrix $G^l$ and $\hat{G}^l$ are calculated, where $G^l, \hat{G}^l$ are the correlations of $f^l, \hat{f}^l$ respectively. Through a back propagation algorithm (red line), the errors are propagated and the  L-BFGS algorithm is used to compute the final result.}
\label{fig:flowchart}
\end{figure*}

\subsection{Texture generation}

As proposed in~\cite{gatys2015texture}, new texture samples are generated by starting from a white noise image and by iteratively imposing the previously defined correlation matrices, using gradient descent. Fig.~\ref{fig:flowchart} shows the flowchart of this algorithm.

Given an image $\hat{I}$ initialized  by white noise, its feature maps $\hat{f}^l$ are computed at each layer of the CNN, as well as the correlation matrix $\hat{G}^l$. A loss function at layer $l$ is then defined as the difference between the correlation matrices of the generated image and the reference one:
\begin{align}
    E_{l} = \frac{1}{4 N_l^2 m_l^2} \sum_{p=1}^{m_l} \sum_{q=1}^{m_l} \| G^l_{p,q} - \hat{G}^l_{p,q} \|^2.
\end{align}
The derivative of this loss function is computed as
\begin{align}
\label{eq:der}
\frac{\partial E_l}{\partial \hat{f}^l_p(i)} =
\frac{1}{N_l^2 m_l^2} \sum_{q=1}^{m_l} f^l_q(i) \cdot (G^l_{p,q} - \hat{G}^l_{p,q}).
\end{align}
Note that this equation does not include a positive part as in~\cite{gatys2015texture}, because the feature maps are always positive after the rectified linear transform.
Combining these layers, a total loss function is defined as
\begin{align}
\mathcal{L}_{cnn}(I, \hat{I}) = \sum_{l=0}^L w_l E_l,
\end{align}
where $w_l$ denotes the weight of loss at layer $l$. With the derivative given by Eq.(\ref{eq:der}), a \emph{back-propagation} (BP) algorithm can be applied to propagate the error from layer $l$ to layer $l-1$, which is illustrated in red line in Fig.\ref{fig:flowchart}. Then, at each layer, the loss (or error) includes two aspects: the derivative of the current layer by Eq.~\ref{eq:der} and the propagated error from the later layer. At any given layer in the network, we can obtain the propagated data error, denoted here by $\Delta_{cnn}$. A solution is then computed using the L-BFGS algorithm~\cite{zhu1994bfgs}, from the initialized image $\hat{I}$ and the back-propagated error $\Delta_{cnn}$.

\section{Texture synthesis by CNNs with spectrum constraint}
\label{sec:synspe}

\begin{figure}[htp!]
\centering
  \includegraphics[width=0.455\linewidth]{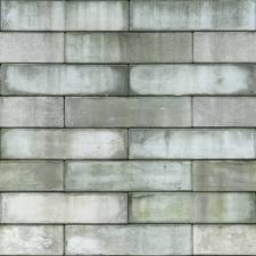}
  \includegraphics[width=0.455\linewidth]{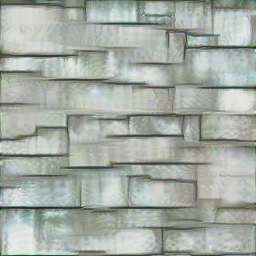}
  	\\\vspace{3px}
  \includegraphics[width=0.455\linewidth]{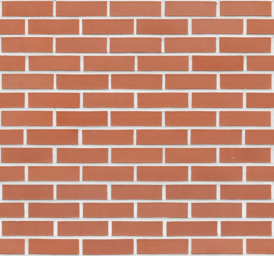}
  \includegraphics[width=0.455\linewidth]{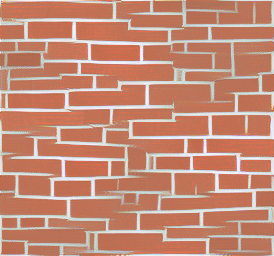}
\caption{Texture synthesis using CNNs according to~\cite{gatys2015texture}. \emph{Left}: a given exemplar texture. \emph{Right}: synthesized result using the CNN model.}
\label{fig:uniquelayer}
\end{figure}

Even though texture synthesis using CNN yields impressive results in many cases, it still suffers from several shortcomings. One of them is the difficulty to accurately reproduce regularity at large scales, typically low-frequency structures. This is illustrated in Fig.\ref{fig:uniquelayer}. In this figure, the left column shows the original images and the right column displays the synthesized images, where large scales are not correctly handled. One possible reason for this is that the size of the convolutional filters in VGG-19 are $3\times 3\times N_l$, which are too small to depict the large structures. Another possible reason is that the number of layers of this network does not allow the handling of very large scale. Nevertheless in our experiments, neither increasing the filter size (this was tested using VGG-m~\cite{Chatfield14} and $7\times 7$ filters) or increasing the number of layers did solve this issue. In this contribution, we show that enriching the loss function of the CNN with a term constraining the Fourier spectrum of the image solves this issue in many cases.

\subsection{Spectrum constraint}
\label{sec:spectrum}

It was shows in~\cite{galerne2011random,galerne2011micro,XiaFPA14} that a large set of textures (the so-called \emph{micro-textures}) may be reproduced simply by imposing the Fourier spectrum of the outputs and letting their Fourier phases be chosen at random. In a different direction, this constraint was used for enabling low frequency structures in sparsity-based texture synthesis~\cite{Tartavel2015}, in a variational framework. We here follow a similar path by constraining the loss function in the CNN-based approach. In what follows, we write $\mathcal{F}(I)$ for the Discrete Fourier Transform (DFT) of an image $I$ and $\mathcal{F}^{-1}$ for the inverse DFT.

\begin{figure}[htp!]
	\centering
	\includegraphics[width=0.46\linewidth]{figure/BrickSmallBrown0293_1_S.png}
	\includegraphics[width=0.46\linewidth]{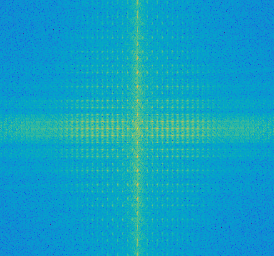}
	\\\vspace{3px}
	\includegraphics[width=0.46\linewidth]{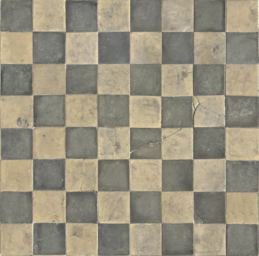}
	\includegraphics[width=0.46\linewidth]{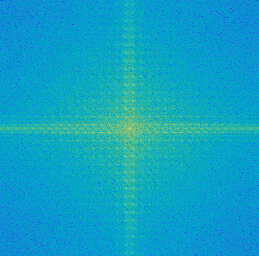}
	\caption{The DFT of (the gray channel of) quasi-periodic images.}
	\label{fig:imgfft}
\end{figure}

In order to illustrate the Fourier representation of low frequency, quasi-periodic structures, Fig.\ref{fig:imgfft} shows the amplitude of the DFT of two images, from which strong peaks can be observed.

Given an image $I$, we write $\mathcal{E}_I$ for the set of images having the same spectrum (modulus of the Fourier transform) as $I$. Simple computations show that the projection of any given image $\hat{I}$ on $\mathcal{E}_I$ (that is, the image from $\mathcal{E}_I$ being closest to $\hat{I}$) is given by
\begin{align}
\label{eq:transferphase}
\tilde{I} = \mathcal{F}^{-1}\left( \frac{\mathcal{F}(\hat{I})\cdot\mathcal{F}(I)^*}{|\mathcal{F}(\hat{I})\cdot\mathcal{F}(I)^*|}\cdot \mathcal{F}(I)  \right),
\end{align}
where $\mathcal{F}(I)^*$ denotes the conjugate of $\mathcal{F}(I)$. For color images, the phase of the gray level image is first computed, and then imposed to each color channel.

\subsection{Texture synthesis by CNNs with spectrum constraint}
In order to constrain both the statistics of the feature maps and the spectrum of the results, we add in the loss function of the CNN an additional term, obtained from the distance of the current image $\hat{I}$ to the space $\mathcal{E}_I$ corresponding to some reference texture $I$. We write $d(\hat{I},\mathcal{E}_I)$ for this term, then its gradient is simply computed as $\hat{I}-\tilde{I}$, where $\tilde{I}$ is computed following Eq.(\ref{eq:transferphase}). Therefore, the gradient term of the CNN is written as
\begin{align}
\label{eq:spe}
\Delta_{spe} =  \hat{I} - \tilde{I},
\end{align}
corresponding to the loss function
\begin{align}
\label{eq:speloss}
  \mathcal{L}_{spe} = \frac{1}{2} d(\hat{I},\mathcal{E}_I)^2,
\end{align}
where as before $I$ is the reference texture.

The final loss function and gradient of the CNN are simply obtained as
\begin{align}
\label{eq:lossall}
\mathcal{L}        = \mathcal{L}_{CNN} + \beta \mathcal{L}_{spe},\\
\label{eq:gradientall}
\Delta = \Delta_{CNN}+ \beta\Delta_{spe}.
\end{align}
The L-BFGS algorithm is then applied to synthesize a new texture image.


\section{Results and analysis}
\label{sec:exp}
\subsection{Experimental setup}
 Following~\cite{gatys2015texture}, the following layers of the convolution neural network are used: 'Conv1\_1', 'Pooling1', 'Pooling2', 'Pooling3', 'Pooling4'. The corresponding weights are set to be $w_1=w_2=w_3=w_4=w_5 = 10^9$. For the spectrum constraint, we use $\beta=10^5$. For the experiments of this paper, we use CG texture samples\footnote{http://www.textures.com} as input. All the images are re-scaled first, as in~\cite{gatys2015texture}.

\subsection{Texture synthesis results}

\begin{figure*}[th]
	\centering
	\includegraphics[width=0.3\linewidth]{figure/FloorsCheckerboard0025_1_S.png}
	\includegraphics[width=0.3\linewidth]{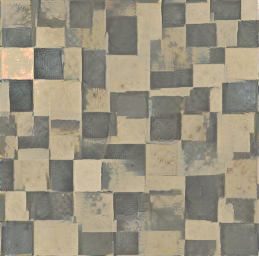}
	\includegraphics[width=0.3\linewidth]{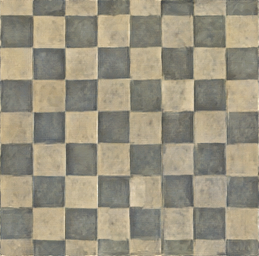}
	\\\vspace{3px}	
	\includegraphics[width=0.3\linewidth]{figure/BrickSmallBrown0293_1_S.png}
	\includegraphics[width=0.3\linewidth]{figure/BrickSmallBrown0293_1_S_syn.png}
	\includegraphics[width=0.3\linewidth]{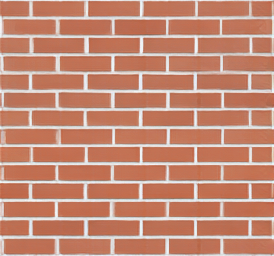}
	\\\vspace{3px}
	\includegraphics[width=0.3\linewidth]{figure/brick3.png}
	\includegraphics[width=0.3\linewidth]{figure/brick3_syn.png}
	\includegraphics[width=0.3\linewidth]{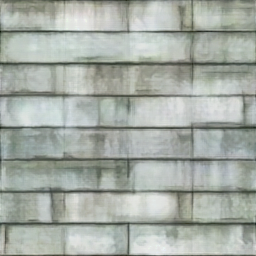}
	\\\vspace{3px}
	\includegraphics[width=0.3\linewidth]{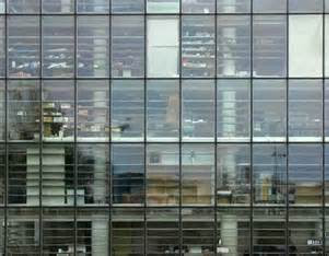}
	\includegraphics[width=0.3\linewidth]{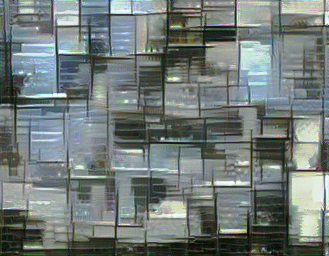}
	\includegraphics[width=0.3\linewidth]{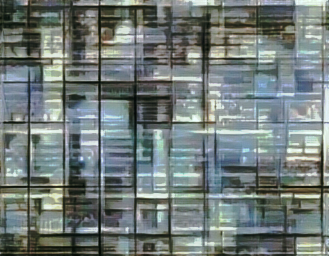}
\caption{\textbf{\emph{Left}}: given texture exemplars;  \emph{\textbf{Middle}}: results obtained by using CNNs as in~\cite{gatys2015texture}; \emph{\textbf{Right}}: our results, obtained by combining the CNN model with spectrum constraints.}
\label{fig:spectrum1}
\end{figure*}

We illustrate the proposed approach by comparing it with the original CNN-based approach from~\cite{gatys2015texture}. This last approach is one of the state-of-the art approaches among methods that produce truly new images. In particular, and contrarily to most patch-based methods derived from~\cite{Efros1999} (see e.g.~\cite{aguerrebere2013exemplar}), these methods (the one from~\cite{gatys2015texture}, as well as the proposed one) do not produce results containing parts that are verbatim copied from the exemplars.

\begin{figure*}[htp!]
	\centering
	\includegraphics[width=0.31\linewidth]{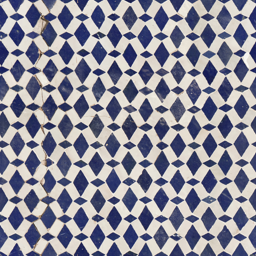}
	\includegraphics[width=0.31\linewidth]{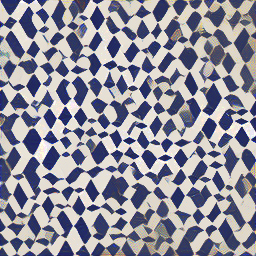}
	\includegraphics[width=0.31\linewidth]{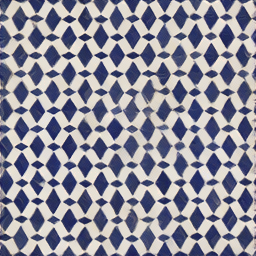}
	\\\vspace{3px}
	\includegraphics[width=0.31\linewidth]{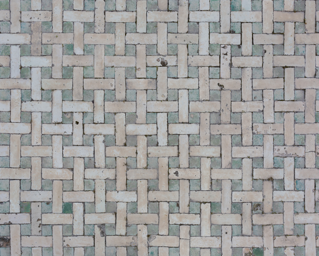}
	\includegraphics[width=0.31\linewidth]{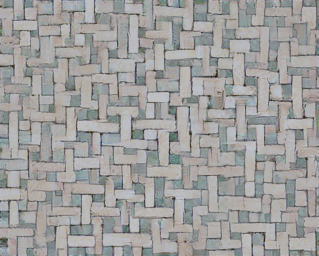}
	\includegraphics[width=0.31\linewidth]{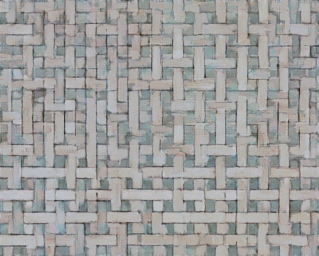}
	\\\vspace{3px}	
	\includegraphics[width=0.31\linewidth]{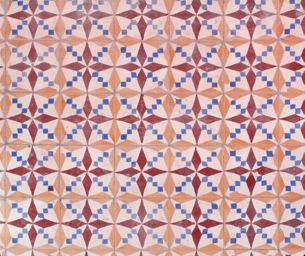}
	\includegraphics[width=0.31\linewidth]{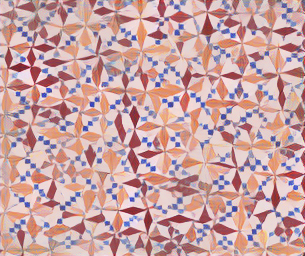}
	\includegraphics[width=0.31\linewidth]{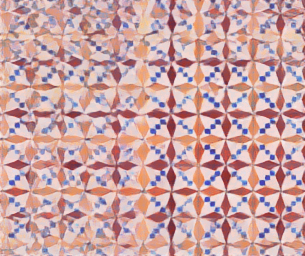}
\caption{\textbf{\emph{Left}}: given texture exemplars;  \emph{\textbf{Middle}}: results obtained by using CNNs as in~\cite{gatys2015texture}; \emph{\textbf{Right}}: our results, obtained by combining the CNN model with spectrum constraints.}
\label{fig:spectrum2}
\end{figure*}

\begin{figure*}[htp!]
	\centering
	\includegraphics[width=0.31\linewidth]{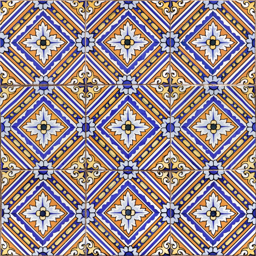}
	\includegraphics[width=0.31\linewidth]{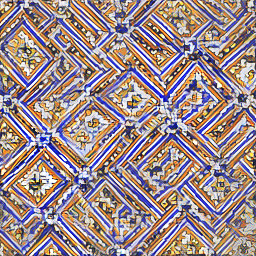}
	\includegraphics[width=0.31\linewidth]{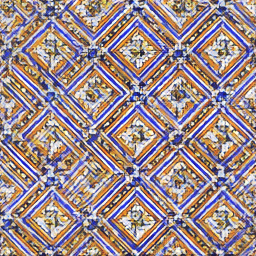}
	\caption{\textbf{A failure example}. Same layout as the previous figures. Observe that though our algorithm (right) can not reproduce the small elements in the exemplar, it still generates a relatively structured result.}
	\label{fig:spectrum3}
\end{figure*}

In Fig.~\ref{fig:spectrum1},~\ref{fig:spectrum2},~\ref{fig:spectrum3}, the left column displays the given texture exemplars, the middle shows the synthesized results using the CNN-based algorithm~\cite{gatys2015texture} and the right illustrates our results by combining the CNN model with a spectrum constraint.

In Fig.~\ref{fig:spectrum1}, from top to bottom are images of a checker board, two brick wall and a building with windows. All these have the particularity to present quasi-periodic structures. In the three first examples, we can see that the algorithm from~\cite{gatys2015texture} fail to preserve the large scale organization, which is correctly reconstructed when adding the spectrum constraint. The last example is more difficult. Although not totally satisfying, the proposed method increases the regularity of the result.

Fig.~\ref{fig:spectrum2} shows synthesis results from Zellige tiles, made of complex decorative patterns. Although some defaults may appear in the global organization of the synthesized images, the combined use of Fourier spectrum constraints and CNN permits the reproduction of both the global patterns and the small scale details. Fig.~\ref{fig:spectrum3} shows an example of failure, where the global structures are too complex to be reproduced by the proposed approach.

Although we do not illustrate it here, it is also worth mentioning that in the cases where the original approach from~\cite{gatys2015texture} is sufficient, results are not degraded by adding the spectrum constraint.

The proposed approach has roughly the same complexity as the one from~\cite{gatys2015texture}, the computation of the FFT being neglectable compared to CNN computations.  Each experiment displayed in this paper took approximately $15$ minutes using a $4$ kernel CPU.

\section{Conclusion}
\label{sec:conclusion}
This paper presents an effective improvement for the synthesis of textures using convolutional neural networks (CNNs), by incorporating a spectrum constraint in the loss function of the CNN. The generated texture images not only preserve local structures and fine scale details, but also preserve large quasi-periodic structures. The experimental results prove that the spectrum constraint is a necessary complement for texture synthesis, especially for structured textures, at no additional computational cost.

\bibliographystyle{ieeetr}
\bibliography{ref}

\begin{thebibliography}{10}

\bibitem{cross1983markov}
G.~R. Cross and A.~K. Jain, ``Markov random field texture models,'' {\em IEEE
  Transactions on Pattern Analysis and Machine Intelligence}, no.~1,
  pp.~25--39, 1983.

\bibitem{julesz1962visual}
B.~Julesz, ``Visual pattern discrimination,'' {\em IRE Transactions on
  Information Theory}, vol.~8, no.~2, pp.~84--92, 1962.

\bibitem{heeger1995pyramid}
D.~J. Heeger and J.~R. Bergen, ``Pyramid-based texture analysis/synthesis,'' in
  {\em Proceedings of the 22nd annual conference on Computer graphics and
  interactive techniques}, pp.~229--238, ACM, 1995.

\bibitem{Portilla2000}
J.~Portilla and E.~P. Simoncelli, ``{Aparametric texture model based on joint
  statistics of complex wavelet coefficients},'' {\em International Journal of
  Computer Vision}, vol.~40, no.~1, pp.~49--71, 2000.

\bibitem{galerne2011random}
B.~Galerne, Y.~Gousseau, and J.-M. Morel, ``Random phase textures: Theory and
  synthesis,'' {\em IEEE Transactions on Image Processing}, vol.~20, no.~1,
  pp.~257--267, 2011.

\bibitem{XiaFPA14}
G.-S. Xia, S.~Ferradans, G.~Peyr{\'{e}}, and J.~Aujol, ``Synthesizing and
  mixing stationary gaussian texture models,'' {\em {SIAM} J. Imaging
  Sciences}, vol.~7, no.~1, pp.~476--508, 2014.

\bibitem{Tartavel2015}
G.~Tartavel, Y.~Gousseau, and G.~Peyr{\'{e}}, ``{Variational Texture Synthesis
  with Sparsity and Spectrum Constraints},'' {\em Journal of Mathematical
  Imaging and Vision}, vol.~52, pp.~124--144, may 2015.

\bibitem{Efros1999}
A.~A. Efros and T.~K. Leung, ``Texture synthesis by non-parametric sampling,''
  {\em The Proceedings of the Seventh IEEE International Conference on Computer
  Vision}, vol.~2, no.~September, pp.~1033--1038, 1999.

\bibitem{Wei2000}
L.-Y. Wei and M.~Levoy, ``{Fast texture synthesis using tree-structured vector
  quantization},'' {\em Proceedings of the 27th annual conference on Computer
  graphics and interactive techniques}, pp.~479--488, 2000.

\bibitem{Lefebvre2005}
S.~Lefebvre and H.~Hoppe, ``{Parallel controllable texture synthesis},'' {\em
  ACM Transactions on Graphics}, vol.~24, no.~3, p.~777, 2005.

\bibitem{aguerrebere2013exemplar}
C.~Aguerrebere, Y.~Gousseau, and G.~Tartavel, ``Exemplar-based texture
  synthesis: the efros-leung algorithm,'' {\em Image Processing On Line},
  vol.~2013, pp.~223--241, 2013.

\bibitem{raad2015conditional}
L.~Raad, A.~Desolneux, and J.-M. Morel, ``Conditional gaussian models for
  texture synthesis,'' in {\em Scale Space and Variational Methods in Computer
  Vision}, pp.~474--485, Springer, 2015.

\bibitem{Dai2014}
J.~Dai, Y.~Lu, and Y.-N. Wu, ``{Generative Modeling of Convolutional Neural
  Networks},'' {\em arxiv}, dec 2014.

\bibitem{gatys2015texture}
L.~Gatys, A.~S. Ecker, and M.~Bethge, ``Texture synthesis using convolutional
  neural networks,'' in {\em Advances in Neural Information Processing
  Systems}, pp.~262--270, 2015.

\bibitem{Lu2016}
Y.~Lu, S.-c. Zhu, and Y.~N. Wu, ``{Exploring Generative Perspective of
  Convolutional Neural Networks by Learning Random Field Models}.'' 2016.

\bibitem{Ulyanov2016}
D.~Ulyanov, V.~Lebedev, A.~Vedaldi, and V.~Lempitsky, ``{Texture Networks:
  Feed-forward Synthesis of Textures and Stylized Images},'' tech. rep., mar
  2016.

\bibitem{galerne2011micro}
B.~Galerne, Y.~Gousseau, and J.-M. Morel, ``Micro-texture synthesis by phase
  randomization,'' {\em Image Processing On Line}, vol.~1, 2011.

\bibitem{simonyan2014very}
K.~Simonyan and A.~Zisserman, ``Very deep convolutional networks for
  large-scale image recognition,'' {\em arXiv preprint arXiv:1409.1556}, 2014.

\bibitem{cimpoi2015deep}
M.~Cimpoi, S.~Maji, and A.~Vedaldi, ``Deep filter banks for texture recognition
  and segmentation,'' in {\em Proceedings of the IEEE Conference on Computer
  Vision and Pattern Recognition}, pp.~3828--3836, 2015.

\bibitem{zhu1994bfgs}
C.~Zhu, R.~H. Byrd, P.~Lu, and J.~Nocedal, ``L-bfgs-b: Fortran subroutines for
  large scale bound constrained optimization,'' {\em Report NAM-11, EECS
  Department, Northwestern University}, 1994.

\bibitem{Chatfield14}
K.~Chatfield, K.~Simonyan, A.~Vedaldi, and A.~Zisserman, ``Return of the devil
  in the details: Delving deep into convolutional nets,'' in {\em British
  Machine Vision Conference}, 2014.

\end{thebibliography}

\end{document}